\documentclass[runningheads]{llncs}
\usepackage{graphicx}
\usepackage{verbatim}
\usepackage[dvipsnames]{xcolor}

\usepackage{hyperref}
\usepackage{booktabs}

\authorrunning{Tim Frommknecht et al.}
\newcommand{\ssl}{Semi-Supervised Learning }

\begin{document}
\title{Augmentation Learning for Semi-Supervised Classification}
%
%
\author{Tim Frommknecht \inst{1}
\and
Pedro Alves Zipf\inst{1}
\and
Quanfu Fan\inst{2}
\and
Nina Shvetsova\inst{1}
\and
Hilde Kuehne\inst{1,2}
}
\institute{Goethe University Frankfurt, Frankfurt, Hesse, Germany \and
MIT-IBM Watson AI Lab, Cambridge, MA, USA}
%
\maketitle              
\begin{abstract}
Recently, a number of new \ssl methods have emerged. 
As the accuracy for ImageNet and similar datasets increased over time, the performance on tasks beyond the classification of natural images is yet to be explored. 
Most \ssl methods rely on a carefully manually designed data augmentation pipeline that is not transferable for learning on images of other domains.
In this work, we propose a \ssl method that automatically selects the most effective data augmentation policy for a particular dataset.
We build upon the Fixmatch method and extend it with meta-learning of augmentations.
The augmentation is learned in additional training before the classification training and makes use of bi-level optimization, to optimize the augmentation policy and maximize accuracy. We evaluate our approach on two domain-specific datasets, containing satellite images and hand-drawn sketches, and obtain state-of-the-art results. We further investigate in an ablation the different parameters relevant for learning augmentation policies and show how policy learning can be used to adapt augmentations to datasets beyond ImageNet.
\end{abstract}

\noindent

\section{Introduction}
Convolutional Neural Networks (CNNs) are widely used in many computer vision applications and achieve state-of-the-art performance in many different tasks across various domains. However, training CNNs requires a large amount of data with proper annotation. Data annotation is often done by humans and thus expensive on a large scale, especially if the labeling requires an educated domain expert (e.g., in the case of medical images). 
Many methods have been proposed to address that problem. One possible solution to avoid an expensive annotation process is utilizing \ssl\cite{assran2021semi,berthelot2019mixmatch,kuo2020featmatch,nassar2021all,sohn2020fixmatch,tian2020makes,zhai2019s4l,berthelot2019remixmatch} that requires labeling for only a small fraction of the data and combines training from labeled data with learning from large-scale unlabeled data. Such training can lead to huge improvements \cite{assran2021semi,berthelot2019mixmatch,nassar2021all,sohn2020fixmatch,tian2020makes,zhai2019s4l,berthelot2019remixmatch}, and unlabeled data can often be gathered at low cost, making \ssl a cheap option to increase performance on datasets with a limited amount of labeled data.

In most modern \ssl methods unlabeled data is utilized by applying an augmentation technique in some form. \cite{assran2021semi,berthelot2019mixmatch,kuo2020featmatch,nassar2021all,sohn2020fixmatch,tian2020makes,zhai2019s4l,berthelot2019remixmatch}. Image augmentation \cite{cubuk2020randaugment,lim2019fast,cubuk2019autoaugment,shorten2019survey} is an effective way to alter images while keeping their class information. However, augmentation can not only alter inessential features of the image but also lead to a loss of relevant information. 
Augmentation that changes the colors of an image can be useful for the task of distinguishing between a tractor and a chair but can be harmful to the task of predicting the condition of a diseased tomato leaf.
Most modern \ssl methods \cite{assran2021semi,berthelot2019mixmatch,kuo2020featmatch,nassar2021all,sohn2020fixmatch,tian2020makes,zhai2019s4l,berthelot2019remixmatch} are mostly evaluated on the ImageNet dataset\cite{deng2009imagenet} or similar natural image datasets like CIFAR-10 \cite{krizhevsky2009learning} and CIFAR-100 \cite{krizhevsky2009learning}, where the classification task is to recognize full objects in real life images. And so far the selection of augmentation has mostly been based on the incentive to improve performance on these datasets. Hence, the used augmentations might not be optimal for all datasets, so the methods might not be able to unfold their full potential on domain-specific datasets with unique properties such as medical images or pencil sketches.

To address this problem, we propose a \ssl method that performs data augmentation policy learning, to optimize the \ssl training for each individual, domain-specific dataset.  Inspired by recent advances in augmentation training in self-supervised learning setups, such as AutoAugment~\cite{cubuk2019autoaugment}, Fast AutoAugment~\cite{lim2019fast}, or DADA~\cite{li2020dada}, we propose a way to integrate the augmentation learning pipeline into a Semi-Supervised setup. We combine FixMatch~\cite{sohn2020fixmatch}, a recently published method for \ssl, with the augmentation learning approach of DADA~\cite{li2020dada}. This is done by building a bi-level-optimization loop around FixMatch which iteratively updates augmentation parameters to maximize classification accuracy in future epochs. The augmentation parameters include a weight determining how frequently the augmentation is applied.  The learned augmentation is then applied during follow-up training. To make the best use of the learned augmentation we propose to add a sharpening step, to further improve our results.  In an experimental evaluation, we compare our method to the original FixMatch approach on two datasets with non-ImageNet-like properties. On both datasets, we can observe an increased accuracy.
In the ablation studies, we show the positive impact of sharpening and investigate the influence of the amount of applied training during the follow-up training. It shows that different datasets require different degrees of augmentation. Finally, we do a quantitative analysis of the impact of single augmentation.

\section{Related Work}

\subsection{\ssl}
\label{sec:rw_ssl}
Recently, many semi-supervised learning methods \cite{assran2021semi,berthelot2019mixmatch,kuo2020featmatch,nassar2021all,sohn2020fixmatch,tian2020makes,zhai2019s4l,berthelot2019remixmatch} emerged implementing different algorithms to set new state-of-the-art results on a variety of challenges and datasets \cite{krizhevsky2009learning,deng2009imagenet}. For this purpose \ssl methods leverage unlabeled data to supplement usual supervised machine learning on labeled samples. This can be especially useful in cases where a huge number of unlabeled data is available but additional labeling exceeds existing resources or is plainly inconvenient. A method that can exploit unlabeled data and produce competitive results in the respective domain can be a big asset for many scientific fields. \cite{su2021realistic} evaluated a variety of \ssl algorithms for robustness on datasets containing unlabeled images of classes not represented in the labeled fraction. Still, in most cases the evaluation of those methods \cite{assran2021semi,berthelot2019mixmatch,kuo2020featmatch,nassar2021all,sohn2020fixmatch,tian2020makes,zhai2019s4l,berthelot2019remixmatch} is done using well established datasets like ImageNet \cite{deng2009imagenet} which mainly include natural images in a common setting, in which the accumulation of data is comparatively easy in contrast to more domain-specific tasks which include more restrictive data. 


\paragraph{Pseudo-labeling} \cite{lee2013pseudo} is a simple and generally applicable method for \ssl that enhances supervised training by additionally utilizing unlabeled samples. This is achieved by iteratively using the already trained network to predict future labels for unlabeled data. To do so the model assigns a probability value to all predictable classes which then is transferred into a pseudo label using a predefined threshold. The created pseudo labels can then be used as targets for a standard supervised loss function. In this sense, pseudo labeling has some similarities to self-training.

\paragraph{Consistency Regularization} \cite{hu2017learning} is used by most state-of-the-art \ssl methods and 
aims to increase the robustness of the trained model to perturbations in different image views of the same class. This usually is accomplished by minimizing the distance between two different instances of the same class or image. In visual terms, the goal is to separate data into clusters of their respective classes. 

Recently many \ssl methods \cite{assran2021semi,berthelot2019mixmatch,kuo2020featmatch,nassar2021all,sohn2020fixmatch,tian2020makes,zhai2019s4l,berthelot2019remixmatch}  have been published and evaluated. One of the state-of-the-art \ssl methods for image classification that has recently been published is FixMatch \cite{sohn2020fixmatch}.
\paragraph{FixMatch} \cite{sohn2020fixmatch} combines pseudo-labeling and consistency regularization in a quite simple yet effective approach. Parallel to conventional supervised training on the labeled fraction of the data, additional training is applied to the unlabeled data. For unlabeled samples, two views of the same image are created. One of the images is only transformed slightly (cropping and horizontal flipping), while the other is strongly augmented using RandAugment~\cite{cubuk2020randaugment}. FixMatch refers to these views as weak and strong augmentation respectively. Next, the model predicts a class for the weakly augmented image. If the confidence of the prediction surpasses a certain threshold, the prediction is transformed into a one-hot pseudo-label. The resulting pseudo-label is then used as the target for the strongly augmented image. The FixMatch approach achieved state-of-the-art results on multiple benchmark datasets, however it was mainly evaluated on ImageNet \cite{deng2009imagenet}, CIFAR-10 \cite{krizhevsky2009learning}, CIFAR-100 \cite{krizhevsky2009learning}, and SVHN \cite{netzer2011reading}. 

\subsection{Augmentation Learning}
Most \ssl methods rely heavily on a data augmentation pipeline. Many different approaches of augmentation have been proposed recently \cite{cubuk2020randaugment,berthelot2019remixmatch,lim2019fast,li2020dada}. FixMatch relies on Randaugment \cite{cubuk2020randaugment} that applies randomly sampled operations from a pool of 15 augmentation operations (like rotation, shearing, and autocontrast). However, the augmentation method and the operations in the pool have been designed and tuned on ImageNet and similar datasets but might not be optimal for other domains.

\paragraph{\underline{D}ifferential \underline{A}utomatic \underline{D}ata \underline{A}ugmentation (DADA)} \cite{li2020dada} is a method for augmentation learning. The method uses bi-Level optimization to deferentially optimize augmentation for supervised image classification on a certain dataset. An augmentation policy is learned during an additional augmentation learning phase prior to the actual model training. An augmentation policy consists of a set of sub-policies of two augmentation operations each. We extend the idea of DADA, which so far has only been applied in fully-supervised settings, and make it usable for the case of \ssl. 

\section{Method}
The main idea of the proposed approach is to learn augmentation policies using bi-level optimization in the first phase and use the learned policies as augmentation for the \ssl method during a follow-up training in the second phase. Note that, while FixMatch is used as a reference, the overall architecture is independent of the semi-supervised learning framework and other methods could be used as well. Unlike DADA, we use the respective \ssl architecture during the augmentation learning phase instead of a supervised one. We start by defining our definition of an augmentation policy before introducing the optimization problem of the augmentation learning phase.

\subsection{Augmentation Policy}

\begin{figure}[tb]
    \centering
    \includegraphics[width=200pt]{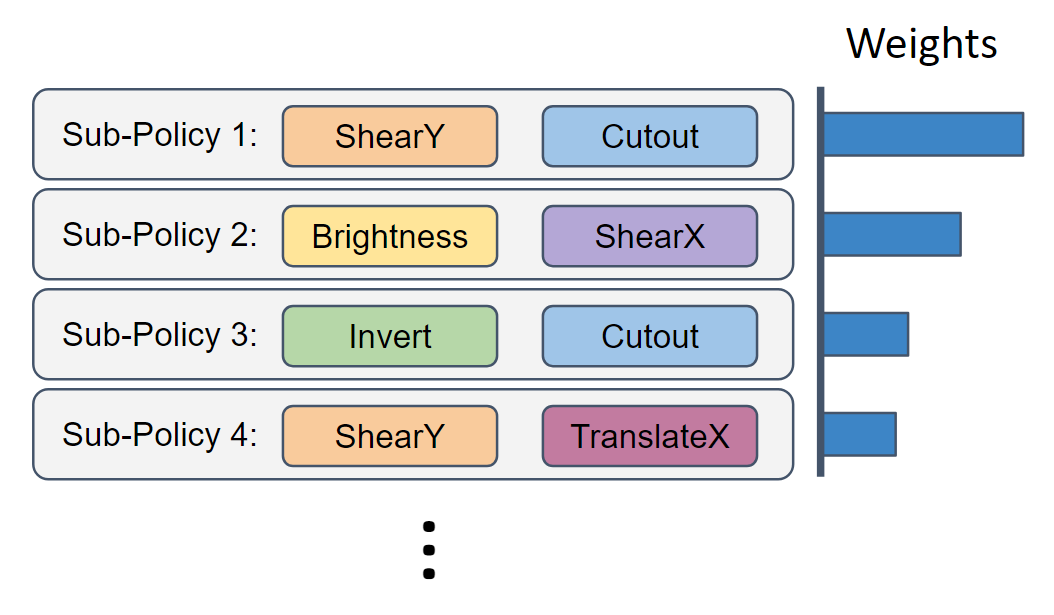}
    \caption{Each sub-policy contains two operations. All sub-policies have a weight. The sub-policies are sampled by a weighted categorical distribution.}
    \label{fig:sub-policies}
\end{figure}

The goal of our system is to find an optimal augmentation policy, for a given dataset. Following Fast AutoAugment \cite{lim2019fast}, we define an augmentation policy $S$ as a set of $N$ sub-policies $s_i$ as described in equation \ref{eq:policy}. 
\begin{equation} 
\label{eq:policy}
    S = \{s_1, s_2, s_3, ..., s_{N} \}
\end{equation}
Where $N$, denotes the total amount of sub-policies. Each sub-policy $s_i$ consists of two operations $O_1$, $O_2$, with respective probabilities $p_1$, $p_2$ and magnitudes $m_1$, $m_2$. The two operations are applied successively. Each operation is applied with its respective probability $p_i$ and magnitude $m_i$. A Bernoulli experiment samples whether an operation is applied using the probability $p_i$. The magnitude of the operation is determined by $m_i$. Most operations are applied with a variable parameter such as rotation angle or intensity. All of these parameters are normalized to a scale from $0$ to $1$. The magnitude parameter $m_i$ lies in the same range and determines the strength of the applied operation. For operations without variable parameters such as invert, the magnitude is simply ignored. Furthermore, each sub-policy is assigned a weight $w$, which represents the chance of the sub-policy being sampled. 
Including all components mentioned above, we define a sub-policy $s_i$ in equation \ref{eq:subpolicy}

\begin{equation} 
\label{eq:subpolicy}
s_i = ((O_{i, 1},\ p_{i, 1},\ m_{i, 1}),\ (O_{i, 2},\ p_{i, 2},\ m_{i, 2}),\ w_i)
\end{equation}
We use 15 different operations, leading to $N=105$ different possible pairs of two operations which we use as sub-policies.
Following DADA \cite{li2020dada}, the selection of sub-policies is modelled as sampling from a categorical distribution with probabilities according to $w$ as visualized in Figure \ref{fig:sub-policies}. The application of operations is modeled as a Bernoulli experiment by the respective probability $p_i$. The search space is the set of all weights $w$, probabilities $p$, and magnitudes $m$ in the augmentation policy $S$


\subsection{Optimization Problem}

\begin{figure}[tb]
    \centering
    \includegraphics[width=\textwidth]{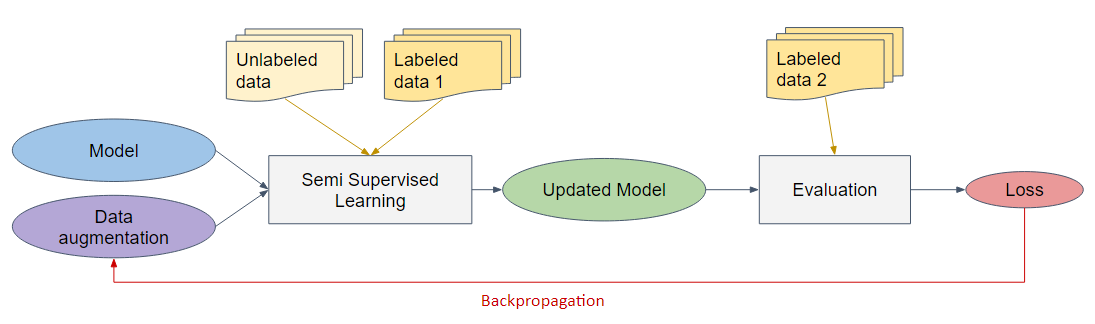}
    \caption{Augmentation learning process: A Data augmentation policy (purple) is used to augment images for a \ssl (FixMatch) update step. Instead of changing the original model (blue), a copy is created (green). All unlabeled data (light yellow) and a part of the labeled data (yellow) are used for the update step. The updated copy (green) is then evaluated using the second part of the labeled data (yellow) resulting in a loss term (red). Finally, the loss is backpropagated through the whole process to compute gradients for the augmentation parameters (purple). For the augmentation learning phase, this architecture is repeated in a training loop parallel to the regular FixMatch training, that updates the model (blue).}
    \label{fig:ssl_dada}
\end{figure}

To find the optimal augmentation policy in the search space described above, we approach an optimization problem as described by DADA \cite{li2020dada}. Our approach builds augmentation learning on top of FixMatch. Parallel to a regular training loop minimizing the loss of the FixMatch classification task, the augmentation training is performed. So, our training consists of two update steps. One to update the model and one to update the augmentation parameters. The two update steps are performed alternately. The FixMatch update step is performed just as described in section \ref{sec:rw_ssl}. After each FixMatch update step, an update step for the augmentation training is applied. One augmentation update step is performed by minimizing a validation loss of a bi-level optimization problem. To do so, we copy the model, sample a sub-policy from a categorical distribution as described above, and perform an update step to the copied model according to the FixMatch loss. The updated copy of the model is then evaluated on an additional fraction of the labeled data, which is only reserved for this particular validation. The resulting validation loss is then differentiated all the way back to the augmentation parameters. A visualization of the augmentation learning process can be seen in Figure \ref{fig:ssl_dada} The gradient flow is estimated by the RELAX gradient estimator \cite{grathwohl2017backpropagation}. For more details see \cite{li2020dada}. For our work, we chose to integrate DADA into FixMatch, but the approach will work for other \ssl methods as well. Once the augmentation training phase is over, the resulting weights $w$ are sharpened. Sharpening is done using the Softmax-function in equation \ref{eq:softmax}.
\begin{equation}
\label{eq:softmax}
    w'_{i} = \frac{\exp({\frac{w_{i}}{T}})}{\sum_{j=1}^{N} \exp({\frac{w_{j}}{T}})}\ \forall i \in \{1, ..., N\}
\end{equation}

The distribution of the weights can be controlled by varying the temperature parameter $T$. A high temperature makes the distribution more uniform, while a low temperature leads to a sharper distribution, s.t. the best sub-policies are used significantly more often than others. A visualization of the effect of sharpening can be seen in Figure \ref{fig:sharpening}.
During the follow-up learning phase, the learned augmentation is used for Fixmatch training. Unlike during the augmentation learning phase, the augmentations are no longer sampled batch-wise but image-wise, s.t. images in the same batch can have different augmentations. Furthermore, the augmentation is no longer limited to one sub-policy at a time. We thereby introduce a parameter $n$ that denotes, how many augmentations are applied to each image.

\paragraph{Integrating DADA augmentation learning into FixMatch:}
For the augmentation training, the FixMatch architecture \cite{sohn2020fixmatch} was adapted to perform the additional augmentation learning step using bi-level optimization after each regular update step. 
Augmentation steps are performed as follows:  For each unsupervised image, two views are created. One weakly augmented one and one strongly augmented one. For weak augmentation, we perform cropping and a random horizontal flip. To perform strong augmentation a single sub-policy is sampled for the whole batch. The resulting unsupervised loss is then used for the bi-level optimization as described above. \\
In parallel, the unsupervised images and an additional fraction of labeled images are used to update the model following the FixMatch method \cite{sohn2020fixmatch}.
For the follow-up training, we apply FixMatch using the learned augmentation as strong augmentation replacing RandAugment \cite{cubuk2020randaugment} from the original approach. 

\section{Experiment}

In this section, we introduce our experimental setup and discuss the performance of our approach. In the first step we generate baseline numbers for our datasets: EuroSAT \cite{helber2017eurosat} and sketch10 \cite{peng2019moment}. Here we use the standard augmentation methods introduced in the original papers respectively. In the second step, we run our FixMatch data augmentation search for each dataset to evaluate the baseline against our augmentation method. This chapter is structured as follows: In section 4.1, we introduce our experimental setup and evaluation methods. In 4.2, the datasets for our experiments are introduced, before we compare our approach to state-of-the-art methods in 4.3. In section 4.4 we'll further investigate our method with an ablation study consisting of quantitative parameter studies and an evaluation of single augmentation operations. 

\subsection{Implementation Details}

\paragraph{Data Splitting:} 
For all experiments, we used 80\% of the data for training and 20\% for testing. While we used all images in the training set, only a small fraction of labels was used. For our experiments, we used 10\% of the training labels. The remaining images are considered unlabeled and the labels are dropped. 
During the augmentation learning phase, we split the labeled images into two partitions of equal size, leading to two labeled datasets containing 5\% of the training data each. While one of the partitions was used as the labeled dataset for the respective \ssl training, the other partition was used for validation in the bi-level optimization during augmentation steps. For the follow-up training, all labeled images (10\% of the train set) were used for \ssl training.

\paragraph{Augmentation learning} For augmentation learning we adjust the setting from DADA for our Semi-Supervised case. We increase the number of trained epochs to 100 and use the dataset as described above. For the learning rate we used $\eta_d = 3 \times 10^{-3}$ and the Adam optimizer. The parameters for the neural network are derived from the official DADA publication \cite{li2020dada} and the batch size is 128. The probabilities $p$ and magnitudes $m$ are uniformly initialized with $0.5$. Meaning that each operation is applied with a $50\%$ chance and with medium magnitude.

\paragraph{Evaluation} All results are reported as average accuracy over all classes on a test set, that was not used for model training or policy search. We compare the proposed approach to the supervised baselines on the labeled images $10\%$, thus using only the labeled part for supervised training, as well as to the fully supervised baseline, which uses the full train set including the labels we dropped for the \ssl experiments. These two can be seen as lower and upper bounds for our experiments. Additionally, we ran these two measures again with RandAugment, to make a fair comparison to the FixMatch algorithm, which uses RandAugment as well.


\paragraph{Network specifications} For all experiments, we use ResNet-50 as the backbone network. We used a pre-trained architecture that was trained on ImageNet. According to the original FixMatch and DADA papers, we used a stochastic gradient descent (SGD) optimizer to update the model and an Adam optimizer to update the augmentation parameters. 

\subsection{Datasets}

\paragraph{EuroSAT} \cite{helber2017eurosat} is an image dataset containing satellite images of different landscapes. The EuroSAT dataset is based on Sentinel-2-Satelite imagery covering 13 spectral bands. To have consistent image formats for better comparison of the augmentations, we choose to use the RGB dataset containing only three of the 13 original channels. The dataset contains 10 different classes with 2000 to 3000 images per class adding up to 27000 total images. Each class represents a certain kind of landscape, like \textit{residential}, \textit{pasture}, or \textit{forest}. We've chosen to use EuroSAT, as its different classes are not recognized by recognizing objects but by environmental structures and properties.

\paragraph{Sketch10} We further derive the Sketch10 dataset as a subset from the original DomainNet dataset \cite{peng2019moment}. To this end, we only use those images of the dataset, that contain data in sketch style. The images are black and white images that contain sketches of 365 different classes. To keep a balanced distribution while at the same time providing enough data, the dataset was further reduced to the 10 largest classes. The reduced dataset contains a total of 6548 images among the 10 classes, with images per class ranging from 593 (\textit{sleeping\_bag}) to 729 (\textit{square}). We propose the Sketch10 dataset as it significantly deviates from conventional object detection datasets, as the dataset contains high-resolution images with a very specific style and features. 

\subsection{Comparison to State-of-the-art}
In this section, we will compare our method of combining FixMatch's \ssl training with DADA's approach of augmentation learning to the original approach FixMatch. We will start by evaluating the baselines using different augmentations, followed by the analysis for FixMatch. The FixMatch analysis begins with an evaluation of the original method to the supervised baselines, followed by a comparison to our method. The following analysis is based on the results in Table \ref{tab:big_table}. 

\paragraph{Supervised baselines} We start by comparing the baselines using RandAugment to the baselines using weak augmentation. We can observe that for Sketch10 RandAugment performs better than weak augmentation. As both baselines (using 10\% and 100\% of data) show a significant improvement when using RandAugment. For EuroSAT, we can observe the opposite behavior, as weak augmentation performs significantly better than RandAugment.  These results indicate, that not every augmentation technique does perform equally well for every dataset, but rather indicate the need for more targeted augmentation learning.

\paragraph{FixMatch VS FixMatch + DADA}
Original FixMatch performs a significant improvement towards the lower bound baselines. For EuroSAT the accuracy is increased by $1.53\%$ and $4.51\%$, compared to the weak and RandAugment baseline respectively. An even stronger improvement can be observed for Sketch10, where the baselines are outperformed by $5.45\%$ (weak) and $4.11\%$ (RandAugment). It further shows that our approach to combining FixMatch with DADA outperforms the original FixMatch in both datasets. For Sketch10 we achieve an improvement of $1.41\%$ in classification accuracy towards the original method. For EuroSAT we could improve performance by $0.42\%$. 
To validate that the improvement is due to the proposed augmentation learning technique and not due to the way the augmentation is applied, we also compare to FixMatch using a random policy, which consists of a set of sub-policies just like our approach. The only difference is, that the weights are defined as uniform and all magnitudes and probabilities are $0.5$. This is done to mimic parameters as they are before training. This way we ensure that the improvement is due to the learning phase.

\begin{table}[tb]
    \centering
    \caption{Comparison of our approach to the original FixMatch and a version of our approach before training, s.t. the policy is not trained but uniformly random. Additionally, we compare our results with supervised baselines using FixMatch's weak augment and Randaugment respectively. Each accuracy was calculated as the average over three runs. The combination of FixMatch and DADA leads to state-of-the-art results for sketch10 and EuroSAT. The Table contains the average accuracy along three runs for each experiment}
    \label{tab:big_table}
    \begin{tabular}{c|c|c|c}
        \toprule 
        & labels used &  Sketch10 & EuroSAT \\
        \midrule
         Supervised baseline (weak augmentation) & 10\% & 61.32 & 93.03\\
         Supervised baseline (RandAugment) & 10\% & 62.66 & 90.05\\
         \midrule
         FixMatch (original) & 10\% & 66.77 & 94.56 \\
         \midrule
         FixMatch + random policy & 10\% & 67.43 & 94.81  \\
         \midrule 
         FixMatch + DADA (ours) & 10\% & \textbf{68.18} & \textbf{94.98} \\
         \midrule 
         \midrule 
         Fully supervised baseline (RandAugment) & 100\% & 75.76 & 95.91\\
         Fully supervised baseline (weak augmentation) & 100\% & 74.03 & 97.09\\
         \bottomrule 
    \end{tabular}
    
\end{table}

\subsection{Ablation and Analysis}
\begin{figure}[tb]
    \centering
    \includegraphics[width=\textwidth]{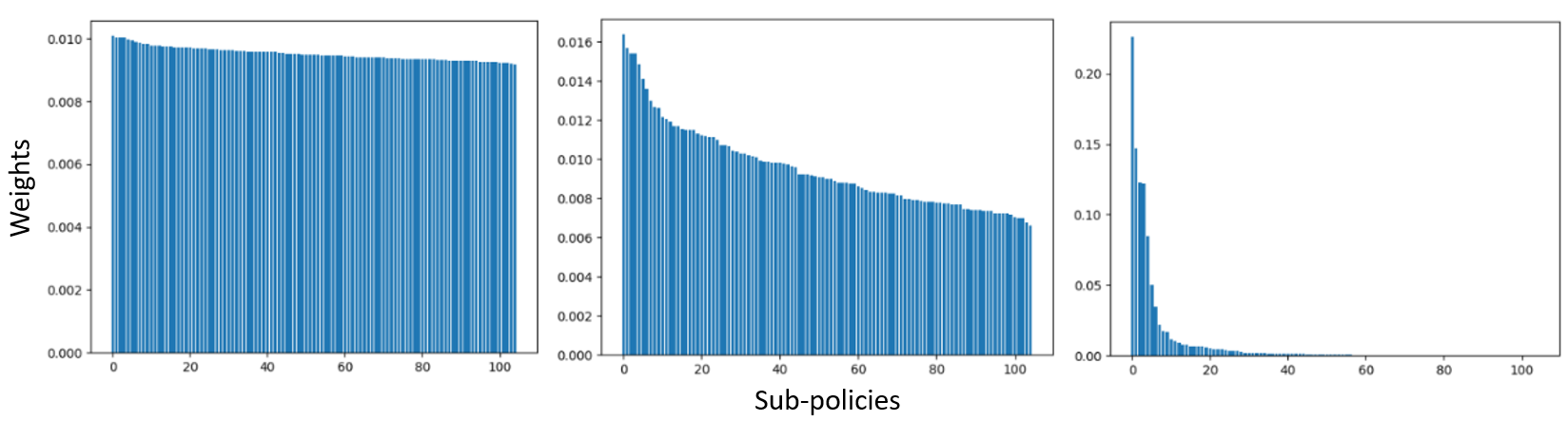}
    \caption{Weight distribution among sub-policies: original (left), sharpened with $T=10^{-3}$ (middle) and sharpened with $T=10^{-4}$ (right). In the figures, each bar represents the weight of a single sub-policy. The sub-policies are ordered from left to right descending by weights. While the original distribution is close to uniform, the sharpened distributions become more selective with sharpening. This effect increases with a lower temperature.}
    \label{fig:sharpening}
\end{figure}

\paragraph{Influence of sharpening Temperature}
One technical contribution of this paper is the sharpening of the learned sub-policy weights after the augmentation training is finished. As described in Section 3.2, sharpening is performed using Softmax with a temperature parameter $T$. Without further sharpening, the weights vary only very little. With different temperatures T, we can control how much the weights deviate. Different levels of sharpness can be seen in Figure \ref{fig:sharpening}. To investigate the influence of $T$ we conduct a parameter study on both datasets. The results and a comparison to uniform sampling ($T=\infty)$ can be seen in Table \ref{tab:T_tuning}. We experimented temperature values ranging from $10^{-4}$ to $10^{-3}$. 

For both datasets, we observe an improvement of the sharpened results towards the original weights. The accuracy varies only very little for EuroSAT but we can still achieve an accuracy gain of $0.09\%$. For Sketch10, we can see a much bigger improvement, as sharpening increases classification accuracy by $2.80\%$. For EuroSAT, the best accuracy is achieved using a lower comparatively low temperature of  $T=1e-4$. For Sketch10, we achieve best accuracy with higher temperatures of $T=9e-4$ and $T=1e-3$. 

\begin{table}[tb]
    \centering
    \caption{Parameter study on EuroSAT for the weight sharpening temperature $T$. The Table contains the average accuracy values over three runs following the FixMatch architecture using our learned augmentation. An infinite temperature ($T=\infty$) leads to a uniform selection of sub-policies. The results are additionally compared to the (unsharpened) original weights and RandAugment. For the runs on EuroSAT, we applied one sub-policy per image during training ($n=1$). For Sketch10 we used $n=4$.}
    \label{tab:T_tuning}
    \begin{tabular}{c|c|c|c|c|c|c|c|c|c|c||c||c}
         \toprule
         T & 1e-4 & 2e-4 & 3e-4 & 4e-4 & 5e-4 & 6e-4 & 7e-4 & 8e-4 & 9e-4 & 1e-3 & $\infty$ & orig. w\\
         \midrule
         EuroSAT & \textbf{94.98} & 94.57 & 94.80 & 94.54 & 94.69 & 94.66 & 94.79 & 94.70 & 94.84 & 94.90 & 94.81 & 94.89 \\
         Sketch10 & 67.47 & 67.16 & 66.34 & 66.97 & 66.44 & 67.74 & 66.26 & 66.50 & \textbf{68.18} & \textbf{68.18} & 67.43 & 65.38 \\
         \bottomrule
    \end{tabular}
    
\end{table}

\paragraph{Quantitative evaluation of the number of applied sub-policies $n$} To explore the behavior of the performance of Fixmatch with harsher augmentation, we apply multiple sub-policies to each image during training. The results can be seen in Table \ref{tab:n_tuning}. The results for both datasets show a clear trend, towards a sweet spot. On EuroSAT, we observe best accuracy with $n=1$, while Sketch10 peaks at $n=4$.  This shows that depending on the dataset a different value for $n$ is optimal which again indicates that augmentation policies can be dataset depended and that it might be desirable to adapt them individually.

\begin{table}[tb]
    \centering
    \caption{Parameter study of FixMatch for the number of applied sub-policies per image $n$. The Table contains the average accuracy along three runs for each experiment. For the experiments, we used a Temperature of $T=9e-4$.} 
    \label{tab:n_tuning}
    \begin{tabular}{c|c|c|c|c}
        \toprule
         n & 1 & 2 & 3 & 4\\
         \midrule
         EuroSAT & \textbf{94.84} & 94.31 & 94.16 & 94.12 \\
         Sketch10 & 66.85 & 67.29 & 65.56 & \textbf{68.18}\\
         \bottomrule
    \end{tabular}
    
\end{table}

\begin{figure}[bt]
    \centering
    \includegraphics[scale=0.28]{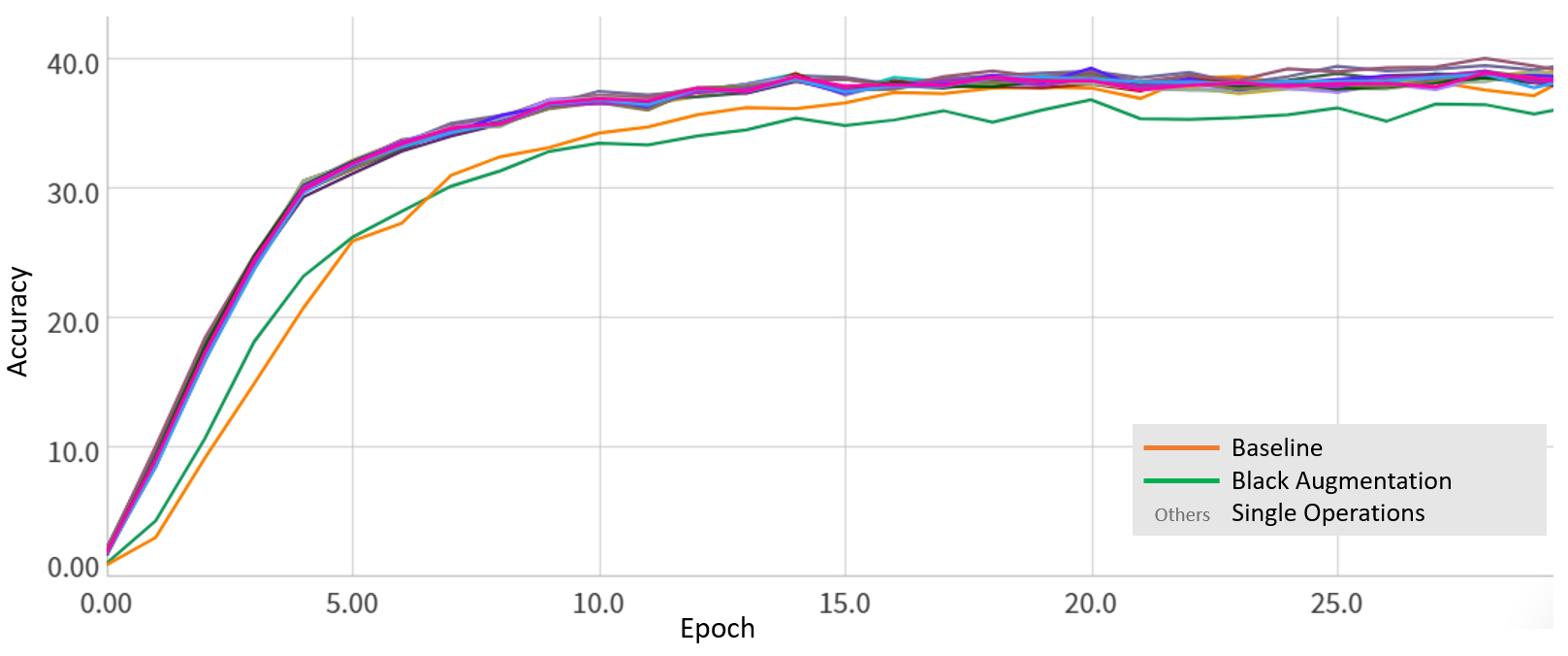}
    \caption{Evaluation of single augmentation operations on Sketch10. Each line represents the accuracy during the first 40 training epochs. In orange, you can see the supervised baseline. The training with the bad augmentation (image set to black) is represented by the green line. Each of the other lines represents an experiment with one of the augmentation operations from our augmentation pool. A selection of the final accuracy values can be seen in table \ref{tab:sketch_single_aug}}
    \label{fig:sketch_single_aug}
\end{figure}

\paragraph{Qualitative evaluation of the influence of single augmentation operations}
Our approach makes use of different augmentation operations by weighting them differently. This way some operations are applied more often, while others are applied less frequently. In Section 4 we've shown, that this leads to an increased accuracy among the tested datasets. To further investigate the influence of single operations we made an additional ablation study on the full Sketch dataset. For these experiments, we took each of our 15 augmentation operations and ran FixMatch training with only that one respective augmentation. So instead of sampling random augmentations from a pool, the same augmentation operation is applied to each image. We use the full Sketch dataset from DomainNet containing all 365 classes. The training process of the experiments can be observed in Figure~\ref{fig:sketch_single_aug}. In addition to the augmentations, we show two baseline experiments for comparison, a supervised baseline with weak augmentation s.t. we can see the improvement caused by the augmentation, as well as one experiment similar to the above but with an augmentation operation, that simply colors the whole image in black. This is supposed to work as a negative example of a bad augmentation, that is expected to mislead training and thus decrease performance. In Figure \ref{fig:sketch_single_aug} we can observe, that the bad augmentation (green) degrades accuracy compared to the supervised baseline (orange). On the contrary, none of our augmentation operations seems to be degrading training as all of them outperform the non-augmented baseline. This leads to the conclusion, that all augmentations can be applied to support training. In Table \ref{tab:sketch_single_aug} we can observe that some augmentations score a higher accuracy than others. This indicates that even though all augmentations do increase accuracy, some operations have a greater effect than others. This supports the idea that all augmentations from the pool can be used but some should be applied more often than others.

\begin{table}[t]
    \centering
    \caption{Accuracy for FixMatch using only one augmentation operation. The table shows the accuracy for a selection of the operations as well as the accuracy for baseline and the purposely bad augmentation. These values refer to the curves in Figure \ref{fig:sketch_single_aug}.  }
    \label{tab:sketch_single_aug}
    \begin{tabular}{c|c|c|c|c|c|c|c}
        \toprule
        Operation & Baseline & Bad & Solarize & Invert & Cutout & Equalize & Color\\
        \midrule
        Accuracy & 38.70 & 36.88 & 39.53 & 38.95 &  38.90 & 40.07 & 38.65\\
        \bottomrule
    \end{tabular}
    
\end{table}

\section{Conclusion}

In this work, we address the problem of learning augmentations in a \ssl setup. While most augmentation methods are designed to perform well on ImageNet or similar datasets, we shift the focus towards more domain-specific datasets such as sketches and satellite images. We propose a novel method for \ssl that trains specific data augmentation for a given dataset. We applied the approach to enhance FixMatch training. We've shown that our new method applies to different datasets and domains, to be precise satellite and sketch images, and has outperformed previous augmentation methods for those settings. Furthermore, we propose to add a sharpening step to the weights of the learned augmentation policy, to further improve the performance of the method and evaluated the effect of different sharpening temperatures in an ablation study. Additionally, we investigated the effect of an increased number of augmentation operations per image. Finally, we investigate the effect of single augmentations for training and find that all augmentations from our pool do improve FixMatch training, but some do more than others. We hope that the proposed method will make \ssl more applicable to satellite or other domain-specific datasets. 

\bibliographystyle{splncs04}
\bibliography{027-main}

\begin{thebibliography}{10}
\providecommand{\url}[1]{\texttt{#1}}
\providecommand{\urlprefix}{URL }
\providecommand{\doi}[1]{https://doi.org/#1}

\bibitem{assran2021semi}
Assran, M., Caron, M., Misra, I., Bojanowski, P., Joulin, A., Ballas, N.,
  Rabbat, M.: Semi-supervised learning of visual features by non-parametrically
  predicting view assignments with support samples. arXiv preprint
  arXiv:2104.13963  (2021)

\bibitem{berthelot2019remixmatch}
Berthelot, D., Carlini, N., Cubuk, E.D., Kurakin, A., Sohn, K., Zhang, H.,
  Raffel, C.: Remixmatch: Semi-supervised learning with distribution alignment
  and augmentation anchoring. arXiv preprint arXiv:1911.09785  (2019)

\bibitem{berthelot2019mixmatch}
Berthelot, D., Carlini, N., Goodfellow, I., Papernot, N., Oliver, A., Raffel,
  C.: Mixmatch: A holistic approach to semi-supervised learning. arXiv preprint
  arXiv:1905.02249  (2019)

\bibitem{cubuk2019autoaugment}
Cubuk, E.D., Zoph, B., Mane, D., Vasudevan, V., Le, Q.V.: Autoaugment: Learning
  augmentation strategies from data. In: Proceedings of the IEEE/CVF Conference
  on Computer Vision and Pattern Recognition. pp. 113--123 (2019)

\bibitem{cubuk2020randaugment}
Cubuk, E.D., Zoph, B., Shlens, J., Le, Q.V.: Randaugment: Practical automated
  data augmentation with a reduced search space. In: Proceedings of the
  IEEE/CVF Conference on Computer Vision and Pattern Recognition Workshops. pp.
  702--703 (2020)

\bibitem{deng2009imagenet}
Deng, J., Dong, W., Socher, R., Li, L.J., Li, K., Fei-Fei, L.: Imagenet: A
  large-scale hierarchical image database. In: 2009 IEEE conference on computer
  vision and pattern recognition. pp. 248--255. Ieee (2009)

\bibitem{grathwohl2017backpropagation}
Grathwohl, W., Choi, D., Wu, Y., Roeder, G., Duvenaud, D.: Backpropagation
  through the void: Optimizing control variates for black-box gradient
  estimation. arXiv preprint arXiv:1711.00123  (2017)

\bibitem{helber2017eurosat}
Helber, P., Bischke, B., Dengel, A., Borth, D.: Eurosat: A novel dataset and
  deep learning benchmark for land use and land cover classification (2017)

\bibitem{hu2017learning}
Hu, W., Miyato, T., Tokui, S., Matsumoto, E., Sugiyama, M.: Learning discrete
  representations via information maximizing self-augmented training. In:
  International conference on machine learning. pp. 1558--1567. PMLR (2017)

\bibitem{krizhevsky2009learning}
Krizhevsky, A., Hinton, G., et~al.: Learning multiple layers of features from
  tiny images  (2009)

\bibitem{kuo2020featmatch}
Kuo, C.W., Ma, C.Y., Huang, J.B., Kira, Z.: Featmatch: Feature-based
  augmentation for semi-supervised learning. In: European Conference on
  Computer Vision. pp. 479--495. Springer (2020)

\bibitem{lee2013pseudo}
Lee, D.H., et~al.: Pseudo-label: The simple and efficient semi-supervised
  learning method for deep neural networks. In: Workshop on challenges in
  representation learning, ICML. vol.~3, p.~896 (2013)

\bibitem{li2020dada}
Li, Y., Hu, G., Wang, Y., Hospedales, T., Robertson, N.M., Yang, Y.: Dada:
  differentiable automatic data augmentation. arXiv preprint arXiv:2003.03780
  (2020)

\bibitem{lim2019fast}
Lim, S., Kim, I., Kim, T., Kim, C., Kim, S.: Fast autoaugment. Advances in
  Neural Information Processing Systems  \textbf{32},  6665--6675 (2019)

\bibitem{nassar2021all}
Nassar, I., Herath, S., Abbasnejad, E., Buntine, W., Haffari, G.: All labels
  are not created equal: Enhancing semi-supervision via label grouping and
  co-training. In: Proceedings of the IEEE/CVF Conference on Computer Vision
  and Pattern Recognition. pp. 7241--7250 (2021)

\bibitem{netzer2011reading}
Netzer, Y., Wang, T., Coates, A., Bissacco, A., Wu, B., Ng, A.Y.: Reading
  digits in natural images with unsupervised feature learning  (2011)

\bibitem{peng2019moment}
Peng, X., Bai, Q., Xia, X., Huang, Z., Saenko, K., Wang, B.: Moment matching
  for multi-source domain adaptation. In: Proceedings of the IEEE International
  Conference on Computer Vision. pp. 1406--1415 (2019)

\bibitem{shorten2019survey}
Shorten, C., Khoshgoftaar, T.M.: A survey on image data augmentation for deep
  learning. Journal of big data  \textbf{6}(1),  1--48 (2019)

\bibitem{sohn2020fixmatch}
Sohn, K., Berthelot, D., Li, C.L., Zhang, Z., Carlini, N., Cubuk, E.D.,
  Kurakin, A., Zhang, H., Raffel, C.: Fixmatch: Simplifying semi-supervised
  learning with consistency and confidence. arXiv preprint arXiv:2001.07685
  (2020)

\bibitem{su2021realistic}
Su, J.C., Cheng, Z., Maji, S.: A realistic evaluation of semi-supervised
  learning for fine-grained classification. In: Proceedings of the IEEE/CVF
  Conference on Computer Vision and Pattern Recognition. pp. 12966--12975
  (2021)

\bibitem{tian2020makes}
Tian, Y., Sun, C., Poole, B., Krishnan, D., Schmid, C., Isola, P.: What makes
  for good views for contrastive learning? arXiv preprint arXiv:2005.10243
  (2020)

\bibitem{zhai2019s4l}
Zhai, X., Oliver, A., Kolesnikov, A., Beyer, L.: S4l: Self-supervised
  semi-supervised learning. In: Proceedings of the IEEE/CVF International
  Conference on Computer Vision. pp. 1476--1485 (2019)

\end{thebibliography}

\end{document}